\documentclass{article}

\usepackage{arxiv}
\usepackage{url}
\usepackage{fancyhdr}
\usepackage[utf8]{inputenc}
\usepackage[T1]{fontenc}
\usepackage{booktabs}
\usepackage{float}
\usepackage{graphicx}
\usepackage{wrapfig}
\usepackage{amsmath}
\usepackage{amssymb}
\usepackage{iftex}
\usepackage{algorithm}
\usepackage{algorithmic}
\usepackage{subcaption}
\usepackage[numbers,sort&compress]{natbib}
\usepackage{times}
\usepackage{helvet}
\usepackage{courier}
\usepackage{xurl}
\usepackage[hidelinks]{hyperref}

\newcommand{\method}{Conquer}
\newcommand{\projectpage}{\url{https://conquer-project.pages.dev/}}

\title{Continual Quadruped Robots Coordination via Semantic Skill Discovery}
\author{
\textbf{Daoqing Wang$^{1,2}$}\thanks{Equal contribution.},
\textbf{Yuchen Xiao$^{1,2}$}\footnotemark[1],
\textbf{Weixuan Huang$^{1,2}$},
\textbf{Zhilong Zhang$^{1,2,3}$},
\\
\textbf{Shenghua Wan$^{1,2}$},
\textbf{Meng Li$^{1}$},
\textbf{Lei Yuan$^{1,2,3}$},
\textbf{Yang Yu$^{1,2,3}$}
\\
$^1$ National Key Laboratory of Novel Software Technology, Nanjing University, Nanjing, China
\\
$^2$ School of Artificial Intelligence, Nanjing University, Nanjing, China
\\
$^3$ Polixir Technologies, Nanjing, China
\\
\texttt{\{wangdq,xiaoyuchen,zhangzl,wansh,yuanl\}@lamda.nju.edu.cn}
\\
\texttt{\{231220091,menson\}@smail.nju.edu.cn},
\texttt{yuy@nju.edu.cn}
\\
Project page: \projectpage
}
\date{}

\begin{document}
\maketitle

\begin{abstract}

Multi-quadruped coordination has attracted increasing attention due to its enhanced payload capacity, broader contact coverage, and improved adaptability to challenging tasks. Existing methods for multi-quadruped manipulation typically focus on predefined or closed task families, often relying on multi-agent reinforcement learning (MARL) to train task-specific coordination policies. However, such methods struggle in open-ended continual learning settings, where tasks arrive sequentially and robots are expected to acquire new coordination skills while reusing previously learned ones without catastrophic forgetting. To address this challenge, we propose Conquer, a semantic skill-library framework that formulates continual multi-quadruped coordination as a retrieve-adapt-update process. First, to accommodate varying team sizes across tasks, we design a team-structured Self-Allies-Goal (SAG) backbone that supports variable-cardinality robot teams by explicitly modeling each robot’s own state, teammate context, and task goal. For each incoming task, Conquer constructs a task-level semantic descriptor from pre-execution information and retrieves a relevant skill from the library for adaptation. After successful execution, Conquer updates the skill library by extracting trajectory-level semantic descriptors and organizing them according to semantic distance, thereby enabling continual skill accumulation and cross-task knowledge transfer. Simulation experiments show that Conquer achieves a final average success rate of 95.6\%, demonstrating strong forward transfer and negligible catastrophic forgetting. Real-world rollouts on Unitree Go2 teams further validate the deployment feasibility of Conquer for practical multi-quadruped coordination. Simulation and real-robot demonstration videos are available at: \url{https://conquer-project.pages.dev/}.
\end{abstract}

\section{Introduction}

Quadruped robots, as a representative class of legged mobile robots, have achieved great progress in recent years due to their mobility and stability in open-ended environments such as transportation, search and rescue, and industrial automation~\cite{zhou2022open, yuan2023survey,quadsurvey1}. Early studies mainly focused on locomotion, enabling robots to track velocity or trajectory commands and traverse complex terrain through model predictive control, reinforcement learning, and sim-to-real transfer~\cite{Tuci2018cooperative, alonso2017multi, ji2021reinforcement-quad}. As robotic applications move toward physically interacting with objects, quadruped loco-manipulation has become an important research direction, requiring robots to interact with objects while maintaining whole-body stability and safe contact~\cite{feng2025learning, jaafar2024mrcap}. Nevertheless, the payload capacity, contact range, and manipulation stability of a single quadruped robot remain limited by the physical platform.

Multi-quadruped coordination~\cite{feng2025learning, jaafar2024mrcap} provides a natural way to overcome these limitations through distributed contact and cooperative control~\cite{Tuci2018cooperative,quadsurvey1}. It has long been studied in multi-robot systems, where multiple robots coordinate their motions and contact forces to interact with objects that may be difficult for a single robot to manipulate~\cite{Tuci2018cooperative, alonso2017multi}. Recent studies further extend this idea to legged platforms, demonstrating collaborative quadrupedal payload manipulation over challenging terrain~\cite{ji2021reinforcement-quad}, multi-quadruped long-horizon pushing~\cite{feng2025learning}, and joint planning and control for object transport~\cite{jaafar2024mrcap}. Combined with multi-agent reinforcement learning methods~\cite{yuan2023survey}, these works show the promise of multi-quadruped coordination for physically demanding tasks, but they are mostly developed for a specific task setting or a bounded task family. In open-ended environments, such cooperative systems are more likely to encounter a stream of tasks where the size of the multi-quadruped team and the environments may change.

Continual reinforcement learning~\cite{pan2025survey, wang2024comprehensive-cl-survey1, khetarpal2022towards-cl-survey2} offers a promising perspective for this problem. Parameter isolation methods~\cite{kessler2022same-owl, wolczyk2022disentangling} learn multi-headed policies to memorize the decision-making knowledge for old tasks. Replay-based methods mitigate forgetting by storing experience via replay buffers~\cite{rolnick2019experience} or generative models~\cite{chen2024stable-distr, hu2025continual-cod}. Retrieval methods~\cite{nasiriany2023learning-sailor, wan2024lotus, guo2025srsa} generate task semantic representations~\cite{shridhar2022cliport, brohan2023saycan, rho2025language} such as user instructions, structured task descriptions, scene observations, and visual-language summaries, and store them for knowledge retrieval. Specifically, task variations in multi-quadruped teams may change the team size and coordination patterns, and sequential adaptation can lead to catastrophic forgetting. This raises a key question: \textbf{can we build a continual multi-quadruped coordination framework that selectively reuses skills from pre-execution semantics, while still learning through reward-driven interaction on the new task?}

To tackle this question, we propose Conquer, a semantic skill retrieval framework with a retrieve-adapt-update workflow for continual multi-quadruped coordination. When a new task arrives, it first builds a semantic descriptor from the pre-execution semantics and retrieves the most relevant executable skill from the skill library. The parameters of the retrieved skill are used to initialize a new skill and are further adapted through multi-agent reinforcement learning~\cite{yu2022mappo}. After training, Conquer summarizes successful trajectories with a VLM-to-embedding pipeline and uses the resulting descriptor to decide whether the current skill should update an existing library entry or be inserted as a new one. This workflow is implemented with a Self-Allies-Goal (SAG) backbone that provides a shared policy interface across variable team sizes~\cite{wang2020action-asn, hu2021updet}, and with lightweight low-rank adapters~\cite{hu2022lora} that store each coordination skill. Experiments on a 14-task Isaac Lab~\cite{isaaclab} benchmark demonstrate that Conquer achieves strong performance, reaching $95.6\%$ final average success rate with strong forward transfer and negligible forgetting. Further ablations confirm the importance of semantic retrieval and LoRA/LocHead skill transfer, and real-world Unitree Go2 rollouts show the deployment feasibility of Conquer through a hierarchical real-robot control stack.

\section{Related Work}

\paragraph{Multi-Agent Reinforcement Learning (MARL)}
aims to train multiple agents to optimize a shared objective. Different from single-agent settings, MARL faces the curse of dimensionality in the joint state-action spaces, caused by the growing number of agents. To overcome this challenge, centralized training and decentralized execution (CTDE) methods transform the complex joint space into low-dimensional agent-wise subspaces, showcasing high efficiency in real-world applications such as autonomous driving and embodied intelligence~\cite{feng2025multi}. Specifically, VDN~\cite{sunehag2017value-vdn} and QMIX~\cite{rashid2020monotonic-qmix} learn value decomposition networks to reduce the value function dimensionality, while MADDPG~\cite{lowe2017multi-maddpg}, COMA~\cite{foerster2018counterfactual-coma}, and MAPPO~\cite{yu2022mappo} learn decentralized actors for execution efficiency. Furthermore, multi-robot manipulation investigates how MARL methods aid physical robot teams to coordinate contact, relative positioning, and object dynamics in transport, towing, carrying, and pushing tasks~\cite{Tuci2018cooperative, ji2021reinforcement-quad, feng2025learning, jaafar2024mrcap}. These works provide the control and learning foundation for multi-quadruped coordination, focusing on improving performance within a fixed task family.

\paragraph{Continual Reinforcement Learning}
studies how models learn a sequence of tasks while balancing stability and plasticity~\cite{pan2025survey, wang2024comprehensive-cl-survey1, khetarpal2022towards-cl-survey2, tang2025mitigating, abbas2023loss-of-plasticity}. Replay-based methods including DISTR~\cite{chen2024stable-distr} and CoD~\cite{hu2025continual-cod} mitigate forgetting by storing experiences via replay buffers or generative models. Parameter isolation methods such as OWL~\cite{kessler2022same-owl} store knowledge for each task with multi-headed networks, while ClonEx-SAC~\cite{wolczyk2022disentangling} combines multi-headed networks and parameter regularization~\cite{kirkpatrick2017overcoming, aljundi2018memory-mas} for better performance. Gradient projection methods such as GEM~\cite{lopezpaz2017gradient} and SGP~\cite{saha2023continual-sgp} adjust update directions to protect prior knowledge. To evaluate forgetting and transfer of different methods, robotic continual learning benchmarks such as Continual World~\cite{wolczyk2021continual-world} provide various task stream settings designed for single-agent or specific scenarios. Continual coordination~\cite{yuan2024macpro, yuan2023macop} is more challenging as task changes may alter not only individual behaviors but also the cooperative structure among agents, such as contact assignment and team-level equilibria.

\paragraph{Skill Discovery}
learns action chunks via temporal behavioral abstractions, which can improve downstream reinforcement learning~\cite{stolle2002learning-option}. Unsupervised skill discovery methods such as DIAYN~\cite{eysenbach2018diversity-diayn} and DADS~\cite{sharma2020dynamics-dads} learn latent skills to promote exploration, and multi-agent skill discovery methods such as ODIS~\cite{zhang2023discovering-odis} and HiSSD~\cite{liu2025learning-hissd} learn common or task-specific coordination patterns for cross-agent transfer. Retrieval-based robot learning methods such as SAILOR~\cite{nasiriany2023learning-sailor}, LOTUS~\cite{wan2024lotus}, and SRSA~\cite{guo2025srsa} reuse skills by constructing and retrieving task semantics such as user instructions, scene observations and vision-language representations. Furthermore, parameter-efficient adapters~\cite{hu2022lora} provide natural tools for grounding and storing skills for skill-library methods~\cite{wang2026lifelong, liu2025psec}. Despite these advances, existing methods often rely on explicit task boundaries or access to old task datasets, and cannot handle varying team sizes in multi-quadruped task streams, limiting their applicability in open-ended environments. Our framework addresses these issues via the semantics-based skill retrieval and a team-structured Self-Allies-Goal (SAG) backbone.

\section{Problem Setting}
We study multi-quadruped coordination tasks under the framework of decentralized partially observable Markov decision processes (Dec-POMDPs)~\cite{oliehoek2016concise}:
\begin{equation}
\mathcal{M}=\langle\mathcal{N},\mathcal{S},\mathcal{A},\Omega,P,O,\mathbf{R},\gamma\rangle,
\end{equation}
where $\mathcal{N}=\{1,\ldots,n\}$ is the robot set, $\mathcal{S}$ is the global state space, $\mathcal{A}=\mathcal{A}^1\times\cdots\times\mathcal{A}^n$ is the joint action space, and $\Omega$ is the local observation space.
The transition, observation, and per-robot reward functions are defined as:
\begin{equation}
P:\mathcal{S}\times\mathcal{A}\rightarrow\Delta(\mathcal{S}),\qquad
O:\mathcal{S}\times\mathcal{N}\rightarrow\Omega,\qquad
R^i:\mathcal{S}\times\mathcal{A}\rightarrow\mathbb{R}.
\end{equation}
At time $t$, robot $i$ receives only $o_t^i=O(s_t,i)$ and outputs $a_t^i\in\mathcal{A}^i$.
Each robot receives a reward composed of a team term and an individual shaping term, $R_t^i=R_t^{\mathrm{team}}+R_t^{i,\mathrm{self}}$, and optimizes the discounted return $\mathbb{E}\left[\sum_{t=0}^{\infty}\gamma^t R_t^i\right]$.

Continual coordination investigates a sequence of tasks
\begin{equation}
\mathcal{Y}=(\mathcal{M}_1,\ldots,\mathcal{M}_m,\ldots),
\end{equation}
where each task may change the number of robots, object geometry, terrain, initial configuration, and target state.
During stage $m$, the learner can interact only with the current task $\mathcal{M}_m$ and has no access to earlier tasks.
After finishing the first $M$ tasks, the system should recover an appropriate executable policy for any task in $\mathcal{Y}_M=\{\mathcal{M}_1,\ldots,\mathcal{M}_M\}$ without retraining old tasks.
Under this protocol, the policy must share its structure across different multi-robot teams, and new-task learning must avoid overwriting previously learned coordination skills.

\section{Approach}

\begin{figure}[t]
\centering
\includegraphics[width=\linewidth]{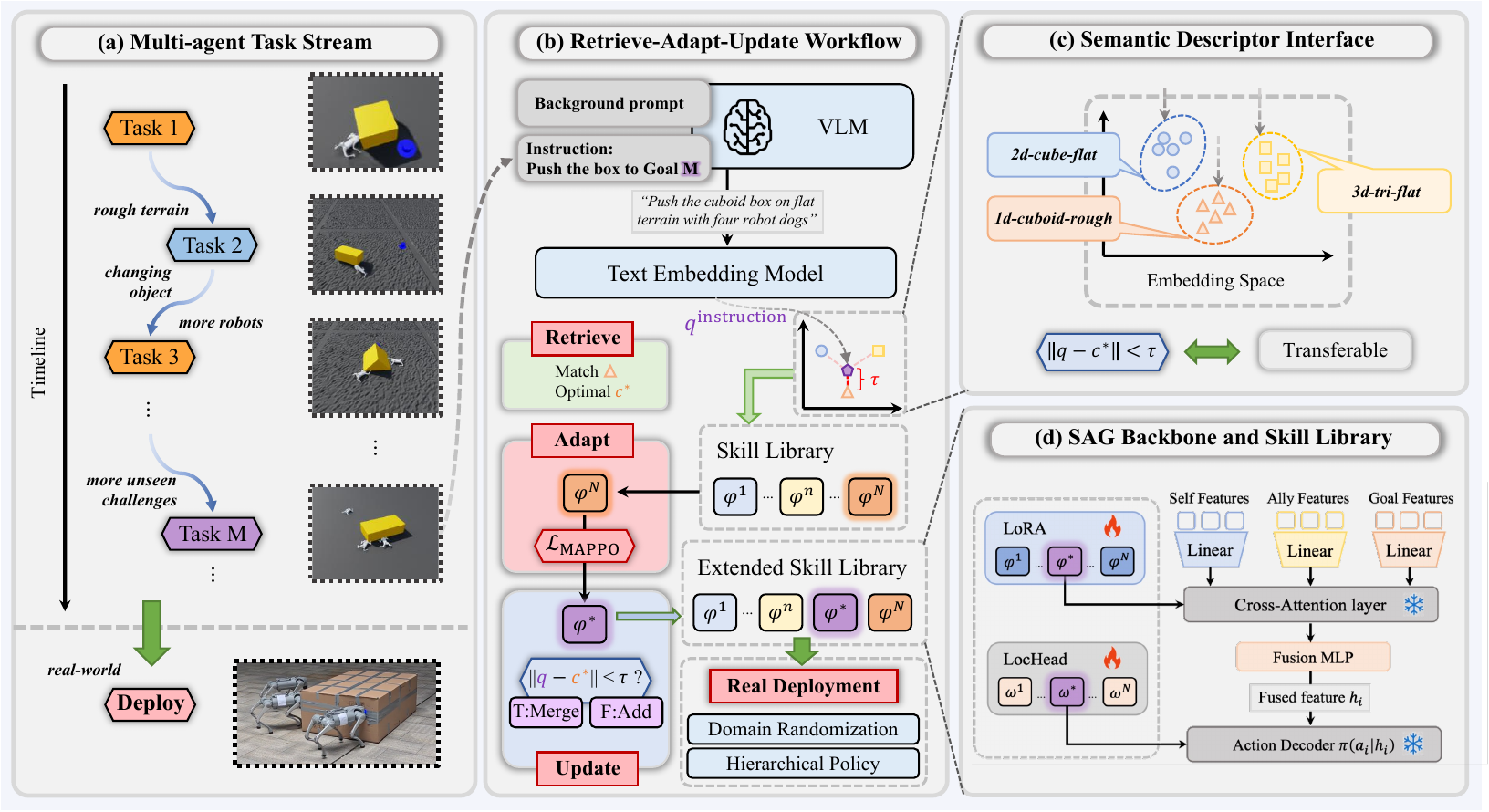}
\caption{Overview of \method{}. (a) Multi-quadruped cooperative tasks arrive sequentially. (b) Each incoming task follows a retrieve-adapt-update workflow that initializes a new skill from a retrieved one, adapts it through MAPPO, and updates the skill library. (c) A VLM-to-embedding interface maps task instructions and trajectories into a shared embedding space for retrieval. (d) A shared SAG backbone is combined with task-specific LoRA/LocHead skill adapters.}
\label{fig:method_overview}
\end{figure}

This section describes the design of \method{} as shown in Figure~\ref{fig:method_overview}.
Sec.~\ref{sec:architecture} introduces the policy architecture and executable skill library.
Sec.~\ref{sec:semantic-descriptor-interface} defines the semantic descriptor interface that maps both incoming task observations and trained skill trajectories into a shared embedding space.
Sec.~\ref{sec:continual-workflow} then organizes these components into a continual retrieve-adapt-update workflow.

\subsection{SAG Backbone and Skill Library}
\label{sec:architecture}

\paragraph{SAG policy backbone.}
\method{} uses a Self-Allies-Goal (SAG) architecture as the shared policy backbone for variable-team coordination.
SAG follows the policy-decoupling idea of UPDeT~\citep{hu2021updet}, and specializes the token structure to multi-quadruped coordination.
For robot $i$, SAG decomposes the local observation into self, ally, and goal tokens, $o^i=[x_i^{\mathrm{self}},x_{i,1}^{\mathrm{ally}},\ldots,x_{i,N_a}^{\mathrm{ally}},x_1^{\mathrm{goal}},\ldots,x_{N_g}^{\mathrm{goal}}]$, and projects them into a shared latent space as $s_i=P_s x_i^{\mathrm{self}}$, $A_i=\{P_a x_{i,k}^{\mathrm{ally}}\}_{k=1}^{N_a}$, and $G=\{P_g x_{\ell}^{\mathrm{goal}}\}_{\ell=1}^{N_g}$.
Using the self token as the query and $C_i=[A_i;G]$ as context, SAG computes
\begin{equation}
\begin{aligned}
u_i =
\mathrm{softmax}\left(\frac{(s_iW^Q)(C_iW^K)^\top}{\sqrt{d_K}}\right)C_iW^V, \quad
h_i = f_{\mathrm{SAG}}(s_i,u_i).
\end{aligned}
\end{equation}
The resulting fixed-dimensional feature $h_i$ is mapped to the continuous action distribution by a lightweight policy head, while the critic can use centralized team information during training.

\paragraph{Skill library.}
Each skill is represented by a lightweight task adapter.
Let $\theta$ denote the frozen SAG policy backbone.
The executable parameters of skill $i$ are $\phi_i=(\phi_i^{\mathrm{lora}},\omega_i^{\mathrm{loc}})$, where $\phi_i^{\mathrm{lora}}$ are LoRA adapter parameters and $\omega_i^{\mathrm{loc}}$ is the LocHead head.
The corresponding policy is $\pi_i(a\mid o)=\pi_{\theta,\phi_i}(a\mid o)$.
At stage $t$, the skill library is $\mathcal{D}_t=\{(s_i,\phi_i,c_i,m_i)\}_{i=1}^{N_t}$, where $s_i$ is the skill identifier, $\phi_i$ is directly loadable for execution or initialization, $c_i$ is a semantic center descriptor, and $m_i$ records environment metadata.
The descriptor $c_i$ is the mean embedding of language descriptions generated from successful trajectories of skill $i$.
In summary, each entry contains both an executable adapter and a semantic index: $\phi_i$ produces grounded control, while $c_i$ supports later retrieval and duplicate detection.
When training a new task, the backbone and old skills stay frozen, while gradients update only the current LoRA/LocHead adapter and the critic.

\subsection{Semantic Descriptor Interface}
\label{sec:semantic-descriptor-interface}

\method{} uses a VLM-to-embedding procedure to construct semantic descriptors for both incoming tasks and stored skills. Given visual observations and a task-level description, a pretrained VLM first produces task-focused text that summarizes core semantics; a text embedding model then maps the text into a shared embedding space $\mathcal{Z}$.
This design follows a common engineering intuition for transfer and curriculum design: tasks with similar team composition, environment context, interaction target, and goal semantics are more likely to share reusable coordination patterns.
We therefore use the Euclidean distance in the descriptor space as a proxy for semantic compatibility: a smaller distance indicates a more plausible source skill for initialization.

For a new task, this procedure transforms a pre-execution frame and task description into an initialization descriptor $q^{\mathrm{init}}$.
For skill $i$, it transforms successful trajectory images into embeddings $Z_i=\{z_{ij}\}_{j=1}^{K_i},z_{ij}\in\mathcal{Z}$.
The skill descriptor is the mean embedding $c_i=\frac{1}{K_i}\sum_{j=1}^{K_i}z_{ij}$.

\subsection{Continual Retrieve-Adapt-Update Workflow}
\label{sec:continual-workflow}

Given the skill library and semantic descriptor interface, \method{} runs a retrieve-adapt-update loop for each incoming task.
In the retrieve step, the pre-execution descriptor selects the nearest stored skill, $i^\star=\arg\min_i\|q^{\mathrm{init}}-c_i\|_2$.
If the library is non-empty, the current LoRA/LocHead adapter is initialized from $\phi_{i^\star}$; otherwise it starts from a fresh adapter.

In the adapt step, the shared SAG backbone and old skills are frozen, while the current adapter $\phi_t$ and critic are optimized for the current task with the MAPPO~\cite{yu2022mappo} objective:
\begin{equation}
\mathcal{L}_{\mathrm{MAPPO}}
=
-\mathbb{E}\left[
\sum_{i\in\mathcal{N}}
\min\left(r_\phi^i A^i,
\mathrm{clip}(r_\phi^i,1-\epsilon,1+\epsilon)A^i\right)
\right]
+\lambda_v\mathcal{L}_V
-\lambda_H\mathcal{H}(\pi_{\theta,\phi_t}).
\label{eq:mappo}
\end{equation}
Here the critic is centralized but agent-indexed: $V_\psi(s_t,i)$ estimates the return for robot $i$ after attending to the team state. $r_\phi^i=\frac{\pi_{\theta,\phi_t}(a^i\mid o^i)}{\pi_{\theta,\phi_{\mathrm{old}}}(a^i\mid o^i)}$ is the PPO ratio for robot $i$.
The advantage is estimated per robot by Generalized Advantage Estimation (GAE)~\cite{schulman2016gae}: $A_t^i=\sum_{\ell\ge0}(\gamma\lambda)^\ell\delta_{t+\ell}^i$ with $\delta_t^i=R_t^i+\gamma V_\psi^i(s_{t+1})-V_\psi^i(s_t)$.
The value loss is averaged over robot-wise value targets, $\mathcal{L}_V=\mathbb{E}_t\!\left[\sum_{i\in\mathcal{N}}\ell_v\!\left(V_\psi^i(s_t),\hat R_t^i\right)\right]$, where $\ell_v$ is the clipped value regression loss used by PPO.
$\mathcal{H}$ is the policy entropy, and $\lambda_v,\lambda_H$ are their weights.

In the update step, successful trajectories from the adapted skill are encoded into $q_t^{\mathrm{skill}}$.
The nearest existing entry is $j^\star=\arg\min_j\|q_t^{\mathrm{skill}}-c_j\|_2$.
If $\|q_t^{\mathrm{skill}}-c_{j^\star}\|_2<\tau$, the matched entry is updated; otherwise, the current LoRA/LocHead adapter and descriptor are inserted as a new skill.
This workflow separates continual learning into semantic initialization, reward-driven adaptation, and controlled library growth. The full procedure is summarized in Appendix~\ref{app:algorithm} and hyperparameter settings are listed in Appendix~\ref{app:algorithm-hyperparameters}.

\section{Experiment}
\label{sec:experiment}

In this section, we evaluate \method{} on continual multi-quadruped pushing tasks in simulation and on real Unitree Go2 deployments.
Our experiments aim to answer: (1) whether \method{} learns new tasks while retaining old skills (Sec.~\ref{sec:main-comparison}); (2) which mechanisms contribute to \method{}'s performance (Sec.~\ref{sec:ablation-case-analysis}); and (3) whether the learned skills transfer to the real-world (Sec.~\ref{sec:real-robot-deployment}).

\subsection{Simulation Experiments}
\label{sec:simulation-experiments}

\subsubsection{Simulation Setup}
\label{sec:simulation-setup}

\begin{figure}[t]
\vspace{-8pt}
\centering
\includegraphics[width=\linewidth]{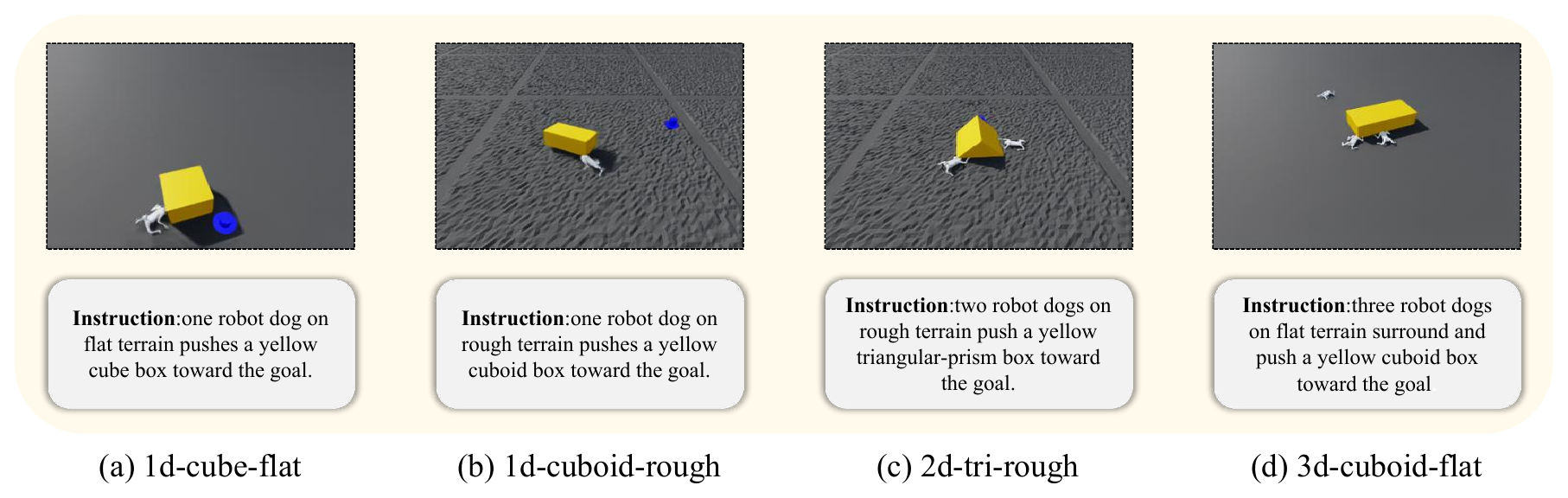}
\vspace{-8pt}
\caption{Representative simulated task views and their retrieval prompts.}
\label{fig:sim_task_sideview_examples}
\vspace{-8pt}
\end{figure}

\paragraph{Benchmark.}
All simulation experiments use Isaac Lab~\citep{isaaclab}.
The benchmark contains 14 canonical push-to-goal tasks that vary robot count, object geometry, and terrain.
The task stream first covers 6 single-dog tasks, then 4 two-dog tasks, and finally 4 three-dog tasks, spanning cube, cuboid, and triangular-prism objects on flat and rough terrains. Figure~\ref{fig:sim_task_sideview_examples} shows examples from this visual-prompt interface.
Detailed task definitions, physical parameters, and success criteria are provided in Appendix~\ref{app:benchmark-details}.

\paragraph{Baselines and Metrics.}
We compare \method{} with online continual-learning baselines including EWC~\citep{kirkpatrick2017overcoming} and Fine-tune, the skill-learning baseline HiSSD and task-aware baseline PSEC. We also include Multitask as a same-training-budget joint-training reference. We report the success rate (SR) for each method and additionally report forward transfer (FWT) and backward transfer (BWT) for online continual-learning methods. Details for the metrics and baselines are described in Appendix~\ref{app:evaluation-protocol} and Appendix~\ref{app:baseline-implementation}.

\subsubsection{Main Results}
\label{sec:main-comparison}

\begin{figure}[t]
\centering
\includegraphics[width=\linewidth]{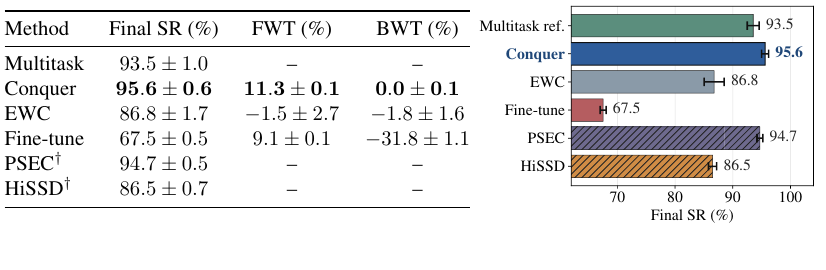}
\makebox[\linewidth][s]{%
    \begin{minipage}[t]{0.56\linewidth}
        \captionsetup{type=table}
        \vspace{-2em}
        \caption{Average metrics $\pm$ standard deviation on the 14-task benchmark across three different seeds. $\dagger$ denotes results obtained under offline-learning protocols.}
        \label{tab:main_results}
    \end{minipage}%
    \hfill
    \begin{minipage}[t]{0.38\linewidth}
    \vspace{-2em}
        \caption{Final success rates. Error bars indicate one standard deviation.}
        \label{fig:main_results_plot}
    \end{minipage}%
}
\vspace{-2em}
\end{figure}

\begin{figure}[t]
\vspace{0.6em}
\centering
\includegraphics[width=1\linewidth]{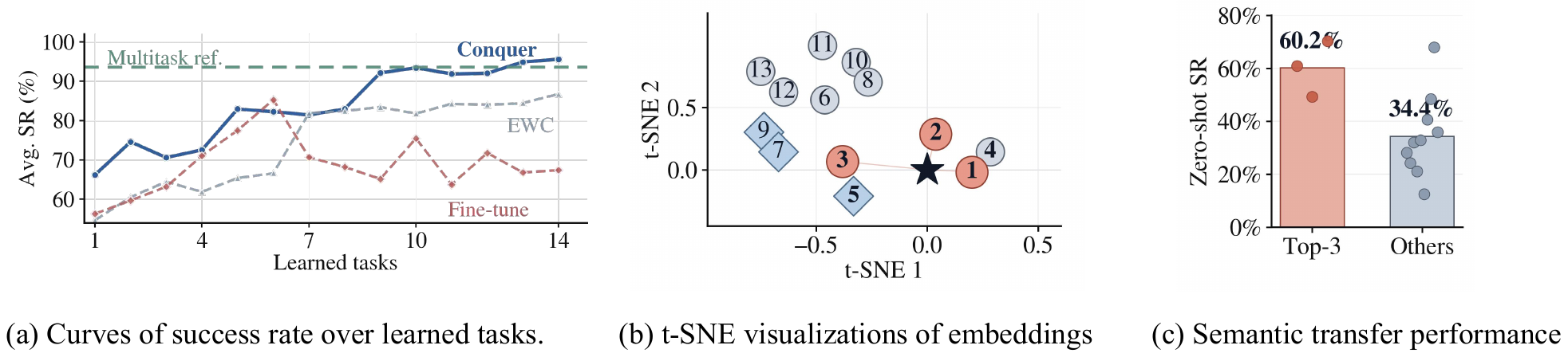}
\vspace{-0.6em}
\caption{Success rate curves and semantic transfer case study for \texttt{3d-tri-rough}.}
\makeatletter
\protected@edef\@currentlabel{\thefigure(a)}\label{fig:growth_curve_methods}
\protected@edef\@currentlabel{\thefigure(b)}\label{fig:semantic_transfer_tsne_3d_tri_rough_b}
\protected@edef\@currentlabel{\thefigure(c)}\label{fig:semantic_transfer_tsne_3d_tri_rough_c}
\makeatother

\vspace{-1.8em}
\end{figure}

We evaluate whether \method{} improves the stability-plasticity trade-off in the continual task stream. As shown in Table~\ref{tab:main_results} and Figure~\ref{fig:main_results_plot}, the basic Fine-tune baseline achieves relatively strong forward transfer, with an FWT of $ 9.1\% $, likely due to the simple-to-harder task order. However, since it lacks protection on previously learned skills when adapting to new tasks, it suffers from severe catastrophic forgetting, leading to its $-31.8\%$ BWT. In contrast, EWC substantially alleviates forgetting by constraining updates to sensitive parameters with the Fisher information matrix, but this comes at the cost of reduced plasticity: it achieves only $ -1.5\% $ FWT, leading to its absolute $8.8\%$ lower final SR compared with our method. Notably, Conquer achieves an FWT of $ 11.3\% $, an approximately zero BWT, and the highest final average SR among online learning baselines under the same training budget, even outperforming the multitask reference in SR. Furthermore, Figure~\ref{fig:growth_curve_methods} shows the evolution of SR of different methods along the task stream, where Fine-tune forgets previous tasks, EWC struggles to adapt to new tasks and Conquer continues to improve its SR and eventually outperforms the multitask baseline.
These results indicate that Conquer achieves the best balance between plasticity and stability among baselines. Detailed semantic-access backtest results are reported in Appendix~\ref{app:semantic-backtest}.

\subsubsection{Ablation and Case Study}
\label{sec:ablation-case-analysis}
To identify the contribution of each component of \method{}, we design three ablation experiments as shown in Table~\ref{tab:ablation_results}, including Random which selects skills randomly at deployment, Multihead which uses only the LocHead as the trainable component, and Scratch which trains each new adapter freshly without initialization from a retrieved one.

\setlength{\columnsep}{2pt}
\setlength{\intextsep}{0pt}
\begin{wraptable}[8]{r}{0.35\linewidth}

\raggedleft
\caption{Ablation results.}
\label{tab:ablation_results}

\begin{tabular}{@{}lc@{}}
\toprule
Condition & Final SR (\%) \\
\midrule
Conquer (Ours) & $95.6 \pm 0.6$ \\
Random & $73.4 \pm 2.5$ \\
Multihead & $86.8 \pm 0.7$ \\
Scratch & $89.2 \pm 0.6$ \\
\bottomrule
\end{tabular}
\vspace{-1.0em}
\end{wraptable}

As shown in Table~\ref{tab:ablation_results}, Random reduces the average success rate by absolute $22.2\%$, indicating that skill selection without task semantics cannot reliably match different tasks. Multihead that removes LoRA causes an absolute $8.8\%$ performance drop, suggesting that task adaptation requires not only output-head selection but also adjustment of intermediate entity-interaction representations. Scratch decreases performance by absolute $6.4\%$, showing that semantic retrieval provides more effective initialization for new tasks. Overall, \method{} achieves the best performance by combining zero-shot skill matching via semantic retrieval with LoRA-based adaptation to different cooperative patterns.

We further present a case study on the \texttt{3d-tri-rough} query to illustrate both the benefit and the limitation of the semantic descriptor in Figure~\ref{fig:semantic_transfer_tsne_3d_tri_rough_b} and Figure~\ref{fig:semantic_transfer_tsne_3d_tri_rough_c}.

In this case, the top-3 semantic neighbors achieve a higher average zero-shot transfer success rate than the remaining candidates. At the same time, semantic rank does not perfectly match physical transferability: the best non-top-3 source, \texttt{2d-tri-flat}, achieves the second-highest transfer success rate among the compared sources despite being outside the top-3 semantic neighbors. Thus semantic retrieval improves expected initialization quality but does not guarantee optimal transfer.
Detailed rank computation, visualization assumptions, and initialization-check values are reported in Appendix~\ref{app:case-study-details}.

\subsection{Real-Robot Deployment}
\label{sec:real-robot-deployment}

To show Conquer's deployment feasibility for real-world multi-quadruped coordination, we evaluate the performance of the learned coordination skills on physical Unitree Go2 teams. We adopt a hierarchical policy architecture for stable real-world deployment: the high-level policy outputs velocity commands $(v_x, v_y, v_{\mathrm{yaw}})$ at 10Hz, while the low-level controller tracks them through motor control at 50Hz. Further deployment details are listed in Appendix~\ref{app:real-robot-details}.

\subsubsection{Real Deployment Setup}
\label{sec:deployment-setup}

\textbf{Sim-to-real policy construction.}
For the one-, two-, and three-Go2 demonstrations, we keep the SAG module trained in simulation and replace the original LocHead with a newly initialized three-dimensional action head that outputs $(v_x,v_y,v_{\mathrm{yaw}})$ for each robot.Meanwhile, the four-Go2 large-cuboid trial uses the same hierarchical action interface, with its policy transferred from the domain-randomized three-Go2 policy.
Then these policies are treated as the high-level policies and are trained through a unified domain-randomization pipeline. We utilize the integrated locomotion policy of the Unitree Go2 as the low-level controller for real-world experiments.

\textbf{Hardware setup.}
Real-world trials are conducted in an outdoor area with an 8.0m $\times$ 8.0m boundary, equipped with a XingYing motion-capture system with 24 cameras, which gathers real-time data regarding robots and objects. The box masses are 1.8kg for the one-robot task, 7.2kg for the two- and three-robot tasks, and 16.2kg for the four-robot task. For each trial, the box is initialized randomly within 3.0--4.5m of a fixed target.

\subsubsection{Real Deployment Results}
\label{sec:deployment-results}

\begin{table}[t]
\vspace{-0.6em}
\centering
\caption{Representative real-robot deployment rollouts.}
\label{tab:real_robot_trials}
\vspace{0.3em}
\begin{tabular}{@{}lccccc@{}}
\toprule
Setting & Dogs & Duration (s) & Start (m) & Final (m) & Min. (m) \\
\midrule
One-Go2 cuboid & 1 & 10  & 3.276  & 0.243  & 0.243  \\
Two-Go2 cuboid & 2 & 20  & 3.365  & 0.800  & 0.800  \\
Three-Go2 cuboid & 3 & 20  & 3.753  & 0.165  & 0.165  \\
Four-Go2 large cuboid & 4 & 20  & 4.457  & 0.304  & 0.059  \\
\bottomrule
\end{tabular}
\vspace{0.3em}
\end{table}

\begin{figure}[t]
\centering
\includegraphics[width=\linewidth]{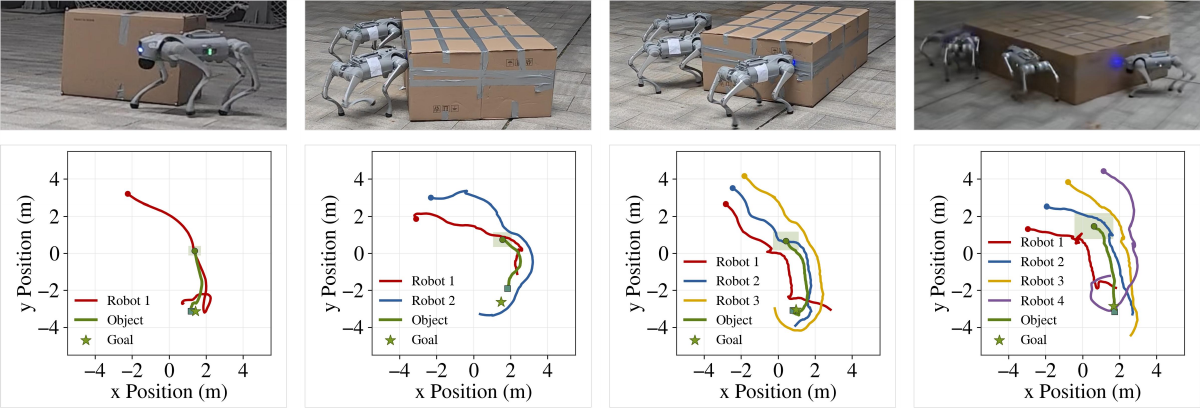}

\caption{Real-robot deployment examples. From left to right, columns show one, two, three, and four Go2 rollout photos with their corresponding trajectories.}
\label{fig:real_robot_deployment_examples}

\end{figure}

Table~\ref{tab:real_robot_trials} reports representative rollouts with their real-world behavior and trajectories. By transferring the policies from simulation to reality, the box-to-goal distances of the one-, three-, and four-Go2 trials reduce from their initial distances to 0.243 m, 0.165 m, and 0.304 m, respectively. The two-Go2 rollout is slower in the fixed 20s window, but the same rollout continues to 0.062m at 29.93s. Concretely, the lower row of Figure~\ref{fig:real_robot_deployment_examples} shows that a single Go2 can flexibly adjust the position of the object until reaching the goal, while 2--4 Go2 coordinate their positioning to push the object, flexibly adjust the direction and movement of the heavier object, and eventually reach the goal. Considering that the real targets and pushed objects have non-negligible physical volume, these results show that the pushing tasks are completed successfully and that the learned coordination policies transfer to hardware in these representative rollouts.

\section{Conclusion}
\label{sec:conclusion}
We present Conquer, a semantic skill-library framework for continual multi-quadruped coordination. Conquer represents learned coordination behaviors through a shared SAG backbone, parameter-isolated adapters, and visual-language descriptors for skill retrieval and library update. By selecting adapters based on pre-execution semantics and adapting skills with MAPPO, Conquer improves new-task learning while keeping old skills accessible. Experiments on a 14-task Isaac Lab benchmark demonstrate strong stability and plasticity, and real-robot rollouts on Unitree Go2 teams further showcase the deployment feasibility of Conquer through a hierarchical control stack.

\paragraph{Limitations and Future Work.} Despite these results, the current study is still limited to a controlled multi-quadruped task family. The retrieval mechanism also relies on the heuristic that nearby semantic descriptors imply reusable coordination behavior, without explicitly modeling physical transferability among environmental dynamics. Future work may extend the static semantic retrieval pipeline toward closed-loop skill management, where training dynamics and physical transfer statistics can be used to revise, merge, or compose candidate skills. Another promising direction is to decompose complex deployment goals into fine-grained cooperative subskills, enabling semantic retrieval to support more heterogeneous multi-robot teams and broader loco-manipulation tasks.

\bibliographystyle{unsrtnat}
\bibliography{references}

\newpage

\appendix

\section{Algorithm and Hyperparameters}
\label{app:algorithm-hyperparameters}
\subsection{Complete Algorithms}
\label{app:algorithm}
\begin{algorithm}[t]
\caption{\method{} retrieve-adapt-update procedure}
\label{alg:conquer}
\small
\begin{algorithmic}[1]
\REQUIRE Task stream $\mathcal{Y}=\{\mathcal{M}_1,\ldots,\mathcal{M}_T\}$; frozen SAG backbone $\theta$; VLM-to-embedding model $E(\cdot)$; descriptor count $K$; duplicate threshold $\tau$; hyperparameters $\mathcal{H}_{\mathrm{rl}},\mathcal{H}_{\mathrm{adapter}}$
\ENSURE Skill library $\mathcal{D}_0\gets\emptyset$
\FOR{$t=1,\ldots,T$}
    \STATE \texttt{// Retrieval}
    \STATE Receive task $\mathcal{M}_t$ and pre-execution input $x_t^{\mathrm{init}}$
    \STATE $q_t^{\mathrm{init}}\gets E(\mathrm{Prompt}_{\mathrm{init}}(x_t^{\mathrm{init}}))$
    \IF{$\mathcal{D}_{t-1}=\emptyset$}
        \STATE Initialize current adapter $\phi_t$ from scratch
    \ELSE
        \STATE $i^\star\gets\arg\min_i\|q_t^{\mathrm{init}}-c_i\|_2$, where $(s_i,\phi_i,c_i,m_i)\in\mathcal{D}_{t-1}$
        \STATE Initialize current adapter $\phi_t\gets\phi_{i^\star}$
    \ENDIF
    \STATE \texttt{// Adaptation}
    \STATE Freeze $\theta$ and all adapters in $\mathcal{D}_{t-1}$
    \STATE Train $\phi_t=(\phi_t^{\mathrm{lora}},\omega_t^{\mathrm{loc}})$ and critic on $\mathcal{M}_t$ with clipped MAPPO
    \STATE \texttt{// Library update}
    \STATE Collect successful  rollouts and generate descriptions $\{d_{t,k}\}_{k=1}^{K}$
    \STATE $Z_t\gets\{E(d_{t,k})\}_{k=1}^{K}$, \quad $c_t\gets\frac{1}{K}\sum_{k=1}^{K}E(d_{t,k})$
    \STATE Record task metadata $m_t$
    \IF{$\mathcal{D}_{t-1}=\emptyset$}
        \STATE $\mathcal{D}_t\gets\{(s_t,\phi_t,c_t,m_t)\}$
    \ELSE
        \STATE $j^\star\gets\arg\min_j\|c_t-c_j\|_2$, \quad $d_t\gets\|c_t-c_{j^\star}\|_2$
        \IF{$d_t<\tau$}
            \STATE $\mathcal{D}_t\gets\mathcal{D}_{t-1}$ with entry $j^\star$ updated by $(\phi_t,c_t,m_t)$
        \ELSE
            \STATE $\mathcal{D}_t\gets\mathcal{D}_{t-1}\cup\{(s_t,\phi_t,c_t,m_t)\}$
        \ENDIF
    \ENDIF
    \STATE Use Algorithm~\ref{alg:conquer_execution} with $\mathcal{D}_t$ to evaluate all tasks and record the growth-curve SR.
\ENDFOR
\STATE \textbf{return} $\mathcal{D}_T$
\end{algorithmic}
\end{algorithm}

\vspace{1.2em}

\begin{algorithm}[t]
\caption{\method{} semantic skill execution and evaluation}
\label{alg:conquer_execution}
\small
\begin{algorithmic}[1]
\REQUIRE Evaluation tasks $\mathcal{Y}^{\mathrm{eval}}=\{\mathcal{M}_1,\ldots,\mathcal{M}_J\}$; skill library $\mathcal{D}_t$; current stage index $t$; frozen SAG backbone $\theta$; VLM-to-embedding model $E(\cdot)$
\item[\textbf{Parameter:}] Number of evaluation episodes $N_{\mathrm{eval}}$
\FOR{$j=1,\ldots,J$}
    \STATE Set up task $\mathcal{M}_j$ and pre-execution input $x_j^{\mathrm{eval}}$
    \STATE $q_j^{\mathrm{eval}}\gets E(\mathrm{Prompt}_{\mathrm{init}}(x_j^{\mathrm{eval}}))$
    \STATE $i^\star\gets\arg\min_i\|q_j^{\mathrm{eval}}-c_i\|_2$, where $(s_i,\phi_i,c_i,m_i)\in\mathcal{D}_t$
    \STATE Load policy $\pi_{\theta,\phi_{i^\star}}$
    \FOR{$p=1,\ldots,N_{\mathrm{eval}}$}
        \STATE Roll out $\pi_{\theta,\phi_{i^\star}}$ on $\mathcal{M}_j$ and record success indicator $u_{j,p}\in\{0,1\}$
    \ENDFOR
    \STATE $R_{t,j}\gets\frac{1}{N_{\mathrm{eval}}}\sum_{p=1}^{N_{\mathrm{eval}}}u_{j,p}$
\ENDFOR
\STATE $G_t\gets\frac{1}{J}\sum_{j=1}^{J}R_{t,j}$
\STATE \textbf{return} success vector $\{R_{t,j}\}_{j=1}^{J}$ and growth-curve SR $G_t$
\end{algorithmic}
\end{algorithm}

To further illustrate the process in Sec.~\ref{sec:continual-workflow}, the overall retrieve-adapt-update workflow is shown in Algorithm~\ref{alg:conquer}: lines 2--10 implement semantic retrieval and adapter initialization, lines 11--13 perform reward-driven adapter training, and lines 14--27 update the skill library with the post-training descriptor.
Line 28 invokes Algorithm~\ref{alg:conquer_execution} after each task stage to evaluate the currently available skill library on the full task set and record the all-task average SR used by the growth curve in Figure~\ref{fig:growth_curve_methods}.
These evaluations do not affect training or library updates.

In Algorithm~\ref{alg:conquer_execution}, evaluation-time semantic routing selects an executable skill for each task, repeated rollouts estimate its task success rate, and the resulting task-wise rates are averaged to obtain the stage-wise all-task SR.
Final SR, FWT, and BWT are computed from the completed task-performance matrix as defined in Appendix~\ref{app:metrics}.

\subsection{Algorithm Hyperparameters}
Table~\ref{tab:appendix_hyperparameters} summarizes the training budget, MAPPO optimizer settings, policy architecture, semantic update rule, and adapter capacity used in the main simulation experiments.
The interaction budget is computed as the number of parallel environments times rollout length times training iterations, yielding the total number of environment steps.
The MAPPO rows specify the inner-loop adaptation optimizer in Eq.~\ref{eq:mappo}, while the SAG and decoder rows define the shared policy backbone used by all skills.
The LoRA rows define the per-skill trainable adapter capacity.

\begin{table}[t]
\centering
\caption{Key settings and interaction budgets used by \method{} in the 14-task simulation benchmark.}
\vspace{8pt}
\label{tab:appendix_hyperparameters}
\begin{tabular}{@{}p{0.60\linewidth}l}
\toprule
Setting & Value \\
\midrule
Interaction budget per 1-/2-robot task & $4096\times48\times500=98.3$M \\
Interaction budget per 3-robot task & $2048\times48\times500=49.2$M \\
PPO epochs & $4$ \\
Learning rate & $3e-4$ \\
value coef $\mathcal{L}_V$ & $1.0$ \\
Entropy coef $\mathcal{L}_H$ & $1e-5$ \\
PPO clipping parameter & $0.2$ \\
Trajectory descriptions per skill $K$ & 16 \\
SAG projection dim & 64 \\
SAG output dim & 64 \\
SAG attention heads & 4 \\
Decoder hidden layer & [256, 128, 64] \\
Duplicate-gate threshold $\tau$ & 0.125 \\
LoRA rank $r$ & 8 \\
LoRA scale $\alpha$ & 16 \\
LoRA dropout & 0.05 \\
Policy total parameters & $165,475$ \\
Trainable policy parameters per skill & $9,379$ \\
\bottomrule
\end{tabular}
\end{table}

\subsection{Vision Language Model, Text embedding model and Prompts}
\label{app:vlm-prompts}
The semantic library uses the same text-embedding space for pre-execution task queries and post-training skill descriptors.
For trajectory-derived skill descriptions, we use \texttt{gemini-robotics-er-1.6-preview} as the vision-language model.
All query and skill-description texts are embedded by \texttt{Qwen/Qwen3-Embedding-0.6B}, which produces 1024-dimensional text embeddings.
Each stored skill descriptor is the mean of $K=16$ description embeddings.

\begin{table}[t]
\centering
\caption{Semantic descriptor components used by \method{}.}
\vspace{8pt}
\label{tab:appendix_vlm_embedding}
\begin{tabular}{@{}p{0.34\linewidth}l}
\toprule
Component & Setting \\
\midrule
Trajectory VLM & \texttt{gemini-robotics-er-1.6-preview} \\
Text embedding model & \texttt{Qwen/Qwen3-Embedding-0.6B}, 1024-dimensional\\
Skill descriptor input & Side oblique view and top bird's-eye view rollouts \\
\bottomrule
\end{tabular}
\end{table}

\begin{table}[t]
\centering
\caption{Prompt templates used to build semantic retrieval queries and stored skill descriptors.}
\vspace{8pt}
\label{tab:appendix_prompt_templates}
\small
\setlength{\tabcolsep}{6pt}
\renewcommand{\arraystretch}{1.16}
\begin{tabular}{|p{0.24\linewidth}|p{0.68\linewidth}|}
\hline
\textbf{Stage} & \textbf{Core Instruction} \\
\hline
Retrieve (pre-adapt) &
Given robot-pushing image(s) and a task instruction, write one short retrieval instruction for the corresponding robot-dog object-pushing skill.
Use a short plain-English sentence that explicitly states the robot count, terrain, object geometry, and intended pushing or moving action.
This is a pre-execution instruction, so do not describe rollout outcomes such as sliding, tilting, or no clear motion.
Use only the provided task/environment facts, preserve the robot count, terrain, and object geometry, and avoid difficulty words unless they appear in the task facts.
Return \texttt{[instruction]} followed by one semi-formal sentence with at most 22 words. \\
\hline
Update (post-adapt) &
Review one timestep from a robot-dog object-pushing rollout.
Use the side oblique view first for object shape and terrain, and use the top bird's-eye view for the number of distinct robot bodies and the contact layout.
Then use a short plain-English sentence that explicitly states the robot count, terrain, object geometry, and intended pushing or moving action.
Return \texttt{[description]} followed by one semi-formal sentence with at most 22 words.\\
\hline
\end{tabular}
\end{table}
\section{Experimental Settings}
\label{app:experimental-settings}

\subsection{Task Stream, Physical Parameters, and Success Criteria}
\label{app:benchmark-details}
The simulation benchmark contains 14 canonical push-to-goal tasks.
The task stream first varies object geometry and terrain for one robot, then increases the team size to two and three robots for heavier cuboid and triangular-prism objects.
Table~\ref{tab:appendix_task_stream} lists the task order and the main physical parameters, and Table~\ref{tab:appendix_task_instructions} reports the corresponding pre-execution retrieval instructions.

\begin{table}[t]
\centering
\caption{Canonical 14-task stream used in simulation. Object sizes are length $\times$ width $\times$ height in meters.}
\vspace{8pt}
\label{tab:appendix_task_stream}
\resizebox{\linewidth}{!}{%
\begin{tabular}{@{}clclccc@{}}
\toprule
Stage & Task ID & Robots & Object & Size (m) & Mass (kg) & Terrain \\
\midrule
1  & \texttt{1d-cube-flat}    & 1 & Cube & $1.0\times1.0\times0.8$ & 5  & Flat \\
2  & \texttt{1d-cube-rough}   & 1 & Cube & $1.0\times1.0\times0.8$ & 5  & Rough \\
3  & \texttt{1d-cuboid-flat}  & 1 & Cuboid & $1.5\times0.8\times0.6$ & 5  & Flat \\
4  & \texttt{1d-cuboid-rough} & 1 & Cuboid & $1.5\times0.8\times0.6$ & 5  & Rough \\
5  & \texttt{1d-tri-flat}     & 1 & Triangular prism & $1.2\times1.2\times0.8$ & 5  & Flat \\
6  & \texttt{1d-tri-rough}    & 1 & Triangular prism & $1.2\times1.2\times0.8$ & 5  & Rough \\
7  & \texttt{2d-cuboid-flat}  & 2 & Cuboid & $2.0\times1.0\times0.6$ & 20 & Flat \\
8  & \texttt{2d-cuboid-rough} & 2 & Cuboid & $2.0\times1.0\times0.6$ & 20 & Rough \\
9  & \texttt{2d-tri-flat}     & 2 & Triangular prism & $1.5\times1.5\times0.8$ & 20 & Flat \\
10 & \texttt{2d-tri-rough}    & 2 & Triangular prism & $1.5\times1.5\times0.8$ & 20 & Rough \\
11 & \texttt{3d-cuboid-flat}  & 3 & Cuboid & $2.4\times1.2\times0.6$ & 30 & Flat \\
12 & \texttt{3d-cuboid-rough} & 3 & Cuboid & $2.4\times1.2\times0.6$ & 30 & Rough \\
13 & \texttt{3d-tri-flat}     & 3 & Triangular prism & $1.8\times1.8\times1.0$ & 30 & Flat \\
14 & \texttt{3d-tri-rough}    & 3 & Triangular prism & $1.8\times1.8\times1.0$ & 30 & Rough \\
\bottomrule
\end{tabular}
}
\end{table}

For each task, the pre-execution semantic query is generated from the task image and the task instruction.
Table~\ref{tab:appendix_task_instructions} lists the canonical instruction sentence used for each environment.

\begin{table}[t]
\centering
\caption{Pre-execution retrieval instructions for the 14 simulation tasks.}
\vspace{8pt}
\label{tab:appendix_task_instructions}
\small
\setlength{\tabcolsep}{5pt}
\renewcommand{\arraystretch}{1.2}
\resizebox{1\linewidth}{!}{
\begin{tabular}{@{}p{0.25\linewidth}l}
\toprule
Task ID & Instruction \\
\midrule
\texttt{1d-cube-flat} & One robot dog on flat terrain pushes a yellow cube box toward the goal. \\
\texttt{1d-cube-rough} & One robot dog on rough terrain pushes a yellow cube box toward the goal. \\
\texttt{1d-cuboid-flat} & One robot dog on flat terrain pushes a yellow cuboid box toward the goal. \\
\texttt{1d-cuboid-rough} & One robot dog on rough terrain pushes a yellow cuboid box toward the goal. \\
\texttt{1d-tri-flat} & One robot dog on flat terrain pushes a yellow triangular-prism box toward the goal. \\
\texttt{1d-tri-rough} & One robot dog on rough terrain pushes a yellow triangular-prism box toward the goal. \\
\texttt{2d-cuboid-flat} & Two robot dogs on flat terrain push a yellow cuboid box toward the goal. \\
\texttt{2d-cuboid-rough} & Two robot dogs on rough terrain push a yellow cuboid box toward the goal. \\
\texttt{2d-tri-flat} & Two robot dogs on flat terrain push a yellow triangular-prism box toward the goal. \\
\texttt{2d-tri-rough} & Two robot dogs on rough terrain push a yellow triangular-prism box toward the goal. \\
\texttt{3d-cuboid-flat} & Three robot dogs on flat terrain surround and push a yellow cuboid box toward the goal. \\
\texttt{3d-cuboid-rough} & Three robot dogs on rough terrain surround and push a yellow cuboid box toward the goal. \\
\texttt{3d-tri-flat} & Three robot dogs on flat terrain surround and push a yellow triangular-prism box toward the goal. \\
\texttt{3d-tri-rough} & Three robot dogs on rough terrain surround and push a yellow triangular-prism box toward the goal. \\
\bottomrule
\end{tabular}
}
\end{table}

\paragraph{Policy observation interface.}
All simulation tasks use the same SAG-structured per-agent observation interface.
For robot $i$, the policy input is organized as
$o_i=[S_i,A_{i,1},\ldots,A_{i,n-1},G_i]$, where $S_i$ is a self token,
$A_{i,j}$ is an ally token for another robot, and $G_i$ is the object-goal token.
The self token contains body-frame base velocity, body-frame angular velocity, projected gravity, joint positions relative to the default pose, joint velocities, local body-object contact information, and terrain height scan.
Each ally token contains the ally robot's relative position, relative velocity, relative orientation, and nearest box-face geometry in the ego robot frame.
The object-goal token contains the box relative pose and velocity, target relative position, box-to-target vector and distance, box size and mass, hold-state indicator, and nearest-face geometry.
The resulting per-agent observation dimensions are 265, 290, and 315 for the one-, two-, and three-robot simulation tasks, respectively.
Table~\ref{tab:appendix_observation_interface} summarizes the token layout.

\begin{table}[t]
\centering
\caption{SAG per-agent observation interface. Token dimensions include internal feedback fields used by the policy but not expanded in the table.}
\vspace{8pt}
\label{tab:appendix_observation_interface}
\begin{tabular}{@{}p{0.13\linewidth}p{0.10\linewidth}p{0.71\linewidth}@{}}
\toprule
Token & Dim. & Contents \\
\midrule
Self & 236 &
\begin{tabular}[t]{@{}l@{}}
Body-frame base velocity (3) \\
Body-frame angular velocity (3) \\
Projected gravity (3) \\
Joint position relative to the default pose (12) \\
Joint velocity (12) \\
Body-object contact force norm (1) \\
Body-object contact normal in the body frame (3) \\
Previous joint-position action command (12) \\
Terrain height scan (187)
\end{tabular} \\
\addlinespace[2pt]
Ally & 25 each &
\begin{tabular}[t]{@{}l@{}}
Ally relative position (3) \\
Ally relative velocity (3) \\
Ally relative orientation quaternion (4) \\
Nearest box-face normal in the ego frame (3)\\
Ally previous joint-position action command (12)
\end{tabular} \\
\addlinespace[2pt]
Object-goal & 29 &
\begin{tabular}[t]{@{}l@{}}
Box relative position (3) \\
Box velocity (3) \\
Box relative orientation quaternion (4) \\
Target relative position (3) \\
Box-to-target vector (3) \\
Planar box-target distance (1) \\
Hold-state indicator (1) \\
Box mass (1) \\
Box half extents (3) \\
Nearest box-face normal (3) \\
Nearest-face signed distance (1) \\
Box angular velocity (3)
\end{tabular} \\
\bottomrule
\end{tabular}
\end{table}

\paragraph{Policy action space.}
In the non-hierarchical simulation benchmark, each robot outputs a 12-dimensional joint-position action, one scalar for each actuated Go2 joint.
The action is applied through a joint-position controller with a scale of 0.25 and the default standing pose as the offset.
The simulator runs at 200Hz with action decimation 4, so the policy acts at 50Hz.

\paragraph{Environment and episode setup.}
All tasks use the official Unitree Go2 USD robot model and a single movable object.
Flat tasks use a planar mesh terrain with static friction 1.0 and dynamic friction 0.95.
Rough tasks use an 8m $\times$ 8m random height-field terrain with 1--6cm height noise and static/dynamic friction 1.0.
Each episode lasts at most 20s, corresponding to 1000 control steps.

\begin{figure}[t]
\centering
\includegraphics[width=\linewidth]{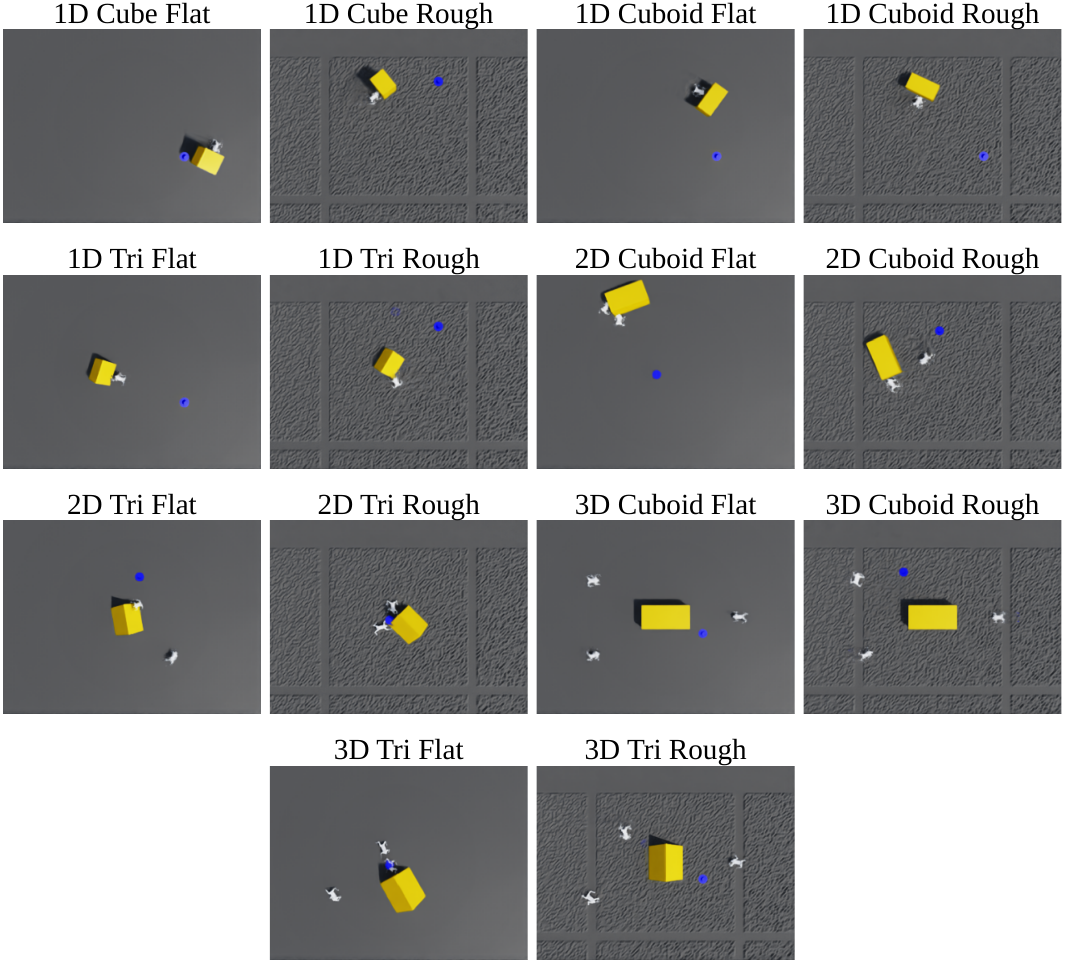}
\caption{Top-view snapshots of the 14 canonical simulation tasks. Panels follow the task order in Table~\ref{tab:appendix_task_stream}.}
\label{fig:appendix_task_topviews}
\end{figure}

\paragraph{Target sampling and success criterion.}
The target command samples a goal point for the object center in the horizontal plane.
For one-robot tasks, the target is sampled on a 3m-radius cylinder around the object.
For two- and three-robot tasks, the target is sampled from the rectangular range $x,y\in[-3,3]$m.
A rollout is counted as successful only after the object center remains within 0.2m of the target in the $xy$ plane for 50 consecutive control steps.
At 50Hz this corresponds to approximately 1s of maintained success.
This maintained-success event, rather than the dense shaping reward, defines the SR metric used in the main text.
Figure~\ref{fig:appendix_task_topviews} provides the corresponding top-view snapshots for the same task stream.

\paragraph{Training rewards.}
The dense reward is used only for reward-driven policy optimization; the reported SR is computed from the maintained-success event described above.
Each robot receives the same reward-term set from the non-hierarchical task factory, and task-level terms are shared by assigning the same object-progress and event rewards to all robots.
Table~\ref{tab:appendix_reward_terms} lists the reward terms used by the 14-task non-hierarchical benchmark.

\begin{table}[t]
\centering
\caption{Reward terms used by the non-hierarchical 14-task simulation benchmark. The reward manager multiplies each raw term by its weight and by the environment time step; event terms internally compensate for this time-step scaling.}
\vspace{8pt}
\label{tab:appendix_reward_terms}
\scriptsize
\setlength{\tabcolsep}{4pt}
\renewcommand{\arraystretch}{1.2}
\resizebox{1\linewidth}{!}{
\begin{tabular}{@{}p{0.29\linewidth}p{0.11\linewidth}p{0.60\linewidth}@{}}
\toprule
Term & Weight & Computation and role \\
\midrule
\texttt{lin\_vel\_z\_l2} & $-2.0$ & Squared vertical base velocity penalty for suppressing jumping. \\
\texttt{ang\_vel\_xy\_l2} & $-0.05$ & Squared roll/pitch angular velocity penalty for stabilizing the base. \\
\texttt{flat\_orientation\_l2} & $-2.0$ & Squared horizontal projected-gravity penalty for keeping the body upright. \\
\texttt{dof\_torques\_l2} & $-2.0{\times}10^{-4}$ & Squared joint-torque penalty for reducing aggressive actuation. \\
\texttt{dof\_acc\_l2} & $-2.5{\times}10^{-7}$ & Squared joint-acceleration penalty for smoother motion. \\
\texttt{action\_rate\_l2} & $-0.01$ & Squared action-difference penalty between consecutive policy outputs. \\
\texttt{feet\_air\_time} & $0.01$ & Foot-air-time reward with a $0.5$s threshold from the contact sensor. \\
\texttt{alive} & $-1.0$ & Constant per-step survival cost, implemented as a unit alive term with negative weight. \\
\texttt{lin\_vel\_heading\_alignment} & $0.5$ & Rewards forward body-frame velocity alignment when the robot moves. \\
\texttt{adaptive\_heading} & $2.5$ & Outside the object coverage radius, rewards velocity aligned with the selected unfinished object. \\
\texttt{adaptive\_activity} & $1.0$ & Inside the selected object's coverage radius, rewards robot planar speed to encourage active pushing contact. \\
\texttt{encourage\_robot\_height} & $1.5$ & Rewards base height up to $0.5$m to encourage standing posture. \\
\texttt{box\_to\_target\_progress} & $20.0$ & Signed object-to-target progress: positive when the object moves closer and negative when it moves away; disabled after maintained success. \\
\texttt{maintain\_box} & $20.0$ & One-time maintained-success event for the object-command pair. \\
\texttt{task\_success} & $10.0$ & Shared terminal success event when the task-success termination is triggered. \\
\texttt{robot\_falling} & $-10.0$ & Shared terminal penalty when any robot-falling termination is triggered. \\
\bottomrule
\end{tabular}
}
\end{table}

This maintained-success event, rather than the dense shaping reward, defines the SR metric used in the main text. Figure~\ref{fig:appendix_task_topviews} provides the corresponding top-view snapshots for the same task stream.

\subsection{Metrics and Baselines}
\subsubsection{Metrics}
\label{app:metrics}
Following the task-performance matrix protocol of GEM~\citep{lopezpaz2017gradient}, we use a task-performance matrix to evaluate the continual-learning process.
Let $T=14$ be the number of tasks and let $R_{i,j}$ denote the success rate on task $j$ after the method has completed training on task $i$.
During evaluation, each task is run with $N_{\mathrm{eval}}$ parallel environments.
If $u_{j,p}\in\{0,1\}$ is the success indicator of the $p$-th rollout on task $j$, then
\begin{equation}
R_{i,j}=\frac{1}{N_{\mathrm{eval}}}\sum_{p=1}^{N_{\mathrm{eval}}}u_{j,p}.
\end{equation}
The success criterion is defined in Appendix~\ref{app:benchmark-details}.
Unless otherwise stated, SR, FWT, and BWT are reported as percentages.

\paragraph{Final success rate.}
Final SR measures the average task coverage after completing the full task stream.
For methods that produce a complete continual-learning matrix, it is the mean of the last matrix row:
\begin{equation}
\mathrm{Final\ SR}=\frac{1}{T}\sum_{j=1}^{T}R_{T,j}.
\end{equation}
For Multitask, PSEC, and HiSSD, which do not follow the setting of online continual learning, we report only the average success rate over the 14 tasks after training.
We therefore do not compute FWT or BWT for these methods.

\paragraph{Forward transfer.}
FWT measures performance on future tasks before those tasks are trained.
Let $b_j$ be the zero-shot success rate of the frozen shared backbone on task $j$.
We compute FWT from the upper-triangular part of the task matrix:
\begin{equation}
\mathrm{FWT}=
\frac{2}{T(T-1)}
\sum_{1\le i<j\le T}
(R_{i,j}-b_j).
\end{equation}
This metric uses only entries where task $j$ has not yet been trained and subtracts the corresponding backbone baseline to reduce the effect of task difficulty.
Higher FWT indicates stronger initialization or retrieval transfer to future tasks.

\paragraph{Backward transfer.}
BWT measures how old-task performance changes after the full sequence is completed:
\begin{equation}
\mathrm{BWT}=
\frac{1}{T-1}
\sum_{i=1}^{T-1}
(R_{T,i}-R_{i,i}).
\end{equation}
Here $R_{i,i}$ is the success rate immediately after learning task $i$, and $R_{T,i}$ is the success rate on the same task after finishing all $T$ tasks.
BWT near zero indicates that old-task performance is maintained, negative BWT indicates forgetting, and positive BWT indicates improvement on old tasks after later training.
Because BWT compares two entries from the same task, no backbone-baseline subtraction is applied.

\paragraph{Growth curve and repeated evaluations.}
The growth curve reports the average success rate after each stage:
\begin{equation}
G_i=\frac{1}{T}\sum_{j=1}^{T}R_{i,j}.
\end{equation}
It is used to visualize the performance trajectory over the task stream, while the main comparison is based on Final SR, FWT, and BWT.
When repeated evaluation runs are available, tables report the mean and sample standard deviation.

\subsubsection{Evaluation-protocol}
\label{app:evaluation-protocol}
For \method{}, continual-learning, and non-continual-learning methods used in this paper, we conducted independent training runs with three different random seeds and evaluated each run in 128 parallel simulation environments, which is equivalent to 128 independent repeated trials. The detailed evaluation procedure is provided in Algorithm~\ref{alg:conquer_execution}.

\subsubsection{Baseline implementations}
\label{app:baseline-implementation}
\textbf{Multitask} is a same-budget joint-training reference inspired by multi-task MARL training protocols~\citep{omidshafiei2017deep-mamt1,wang2023multi-mamt2,Meng2026MTRL}.
Since Isaac Lab evaluates one registered task instance per training process, we implement Multitask as task-rotating joint training: one shared policy is trained over 70 rounds, and each round trains one task for 100 iterations before passing the checkpoint to the next round.
Each of the 14 tasks appears in five rounds, giving 500 iterations per task, matching the per-task budget used by the continual methods.
This reference does not use LoRA skill adapters or semantic retrieval.

\textbf{Fine-tune} is the sequential full-parameter baseline.
It follows the same task order as \method{}, initializes each new task from the previous task checkpoint, and updates all policy parameters with the same clipped MAPPO optimizer~\citep{yu2022mappo}.
LoRA adapters, semantic retrieval, replay, and regularization are disabled, so Fine-tune measures the effect of unconstrained sequential adaptation.

\textbf{EWC} extends the same sequential full-parameter setting with elastic weight consolidation~\citep{kirkpatrick2017overcoming}.
After each task, we estimate a diagonal empirical Fisher matrix from rollout samples and store the task-ending parameter vector.
For later tasks, the MAPPO objective is augmented with a quadratic penalty weighted by the accumulated Fisher estimate.
We use the online form of EWC with Fisher decay, so earlier tasks remain represented in the regularizer while the method continues to adapt to the current task.

\textbf{PSEC} is included as a task-aware parameter-space skill composition reference~\citep{liu2025psec}.
Our adapted implementation builds parameter-space compositions from trained LoRA skill primitives and uses task information to select or combine skill parameters at evaluation time.
Because its task-aware composition protocol does not define a stage-wise, task-agnostic evaluation interface required for FWT/BWT, we report only its final average SR over the 14 tasks.

\textbf{HiSSD} is included as a skill-discovery and hierarchical distillation baseline~\citep{liu2025learning-hissd}.
We use the adapted implementation for this benchmark, which trains a hierarchical skill representation and shared policy from the 14-task data and is evaluated with the same final-step success criterion as the other methods.
Since this offline distillation protocol does not generate after-each-task checkpoints along the online task stream, we report its final average SR only.

\subsection{Real-Robot Deployment Details}
\label{app:real-robot-details}

Sec.~\ref{sec:deployment-setup} reports the hardware platform, motion-capture setup, object masses, and rollout-level results. This section additionally provides the details of simulation-side adaptation and command-interface details used before deployment.

\paragraph{Domain-randomization setting.}
Before deployment, the hierarchical policies are adapted in simulation with randomized contact dynamics, object properties, external perturbations, and command-interface effects. All deployment adaptation tasks use the high-friction box-contact setting. Table~\ref{tab:real_robot_dr} summarizes the randomization factors.

\begin{table}[t]
\centering
\caption{Domain randomization used for real-robot deployment adaptation.}
\vspace{8pt}
\label{tab:real_robot_dr}
\renewcommand{\arraystretch}{1.1}
\resizebox{\linewidth}{!}{
\begin{tabular}{@{}p{0.3\linewidth}p{0.39\linewidth}p{0.35\linewidth}@{}}
\toprule
Factor & Randomization & Notes \\
\midrule
Robot material &
Static friction $[0.7,1.0]$, dynamic friction $[0.5,0.8]$ &
Sampled at startup \\

Box material &
$(1.4,1.35)$, $(1.2,1.15)$, or $(1.0,0.95)$ &
Static/dynamic friction pair \\

Box mass &
Multiplicative scale $[0.7,1.3]$ &
Sampled at reset \\

Robot base mass &
Additive offset $[-1,3]\mathrm{kg}$ &
Sampled at startup \\

External push &
Planar velocity perturbation $v_x,v_y\in[-0.5,0.5]\mathrm{m/s}$ &
Every 10--15s \\

Command delay &
1--6 low-level locomotion ticks &
About 20--120ms at 50Hz \\

Base linear velocity obs. &
Gaussian noise, $\sigma=0.0025\mathrm{m/s}$ &
Low-level policy observation \\

Base angular velocity obs. &
Gaussian noise, $\sigma=0.005\mathrm{rad/s}$ &
Low-level policy observation \\

Projected gravity obs. &
Gaussian noise, $\sigma=0.001$ &
Low-level policy observation \\

Joint position obs. &
Gaussian noise, $\sigma=0.0005\mathrm{rad}$ &
Low-level policy observation \\

Joint velocity obs. &
Gaussian noise, $\sigma=0.005\mathrm{rad/s}$ &
Low-level policy observation \\

Velocity gain &
Scale $[0.7,1.0]$ &
Per-axis, per-episode \\

Velocity bias &
Gaussian bias, $\sigma=0.03\mathrm{m/s}$ &
Per-axis, per-episode \\

Command deadzone &
$[0,0.1]\mathrm{m/s}$ &
Per-episode \\

Command lag &
$\alpha\in[0.3,1.0]$ &
First-order lag \\

Yaw coupling &
$[-0.1,0.1]$ &
Lateral-to-yaw coupling \\
\bottomrule
\end{tabular}
}
\end{table}

The one-robot cuboid adaptation uses a real-size cuboid proxy of $0.714\mathrm{m}\times0.524\mathrm{m}\times0.288\mathrm{m}$ with a 5kg nominal simulation mass. The two- and three-robot recipes use a $1.40\mathrm{m}\times1.00\mathrm{m}\times0.50\mathrm{m}$ cuboid with 20kg and 30kg nominal simulation masses, respectively. The four-robot large-cuboid recipe uses a $2.10\mathrm{m}\times1.40\mathrm{m}\times0.50\mathrm{m}$ cuboid with a 60kg nominal simulation mass. Multi-robot deployment adaptation uses ring-style robot initialization, target sampling around the box with an 8m cylinder radius, and reset-time yaw randomization for both robots and the box.

\paragraph{Deployment model and command waveforms.}
The deployed controller is a high-level decision-layer policy. It maps motion-capture-based observations to per-robot body-frame velocity commands, while the onboard Unitree Go2 controller tracks these commands at the locomotion level (for deployment, motion-capture states and onboard Go2 proprioceptive measurements are converted into the corresponding ego-frame SAG observations, as detailed in Appendix~\ref{app:benchmark-details}). The command-waveform plots report the post-clamp commands streamed to hardware. The conservative command envelope is
$v_x\in[-0.4,0.75]$m/s, $v_y\in[-0.25,0.25]$m/s, and
$v_{\mathrm{yaw}}\in[-0.5,0.5]$rad/s. Saturated waveform segments therefore indicate the deployed safety clamp rather than an unconstrained policy output.

\begin{figure}[t]
\centering
\begin{minipage}[t]{0.48\linewidth}
\centering
\includegraphics[width=\linewidth]{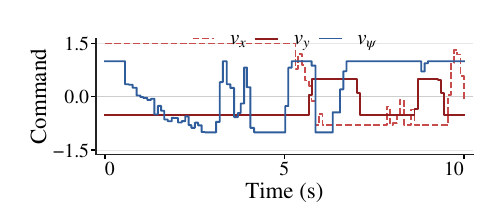}
{\footnotesize (a) One-Go2 cuboid}
\end{minipage}\hfill
\begin{minipage}[t]{0.48\linewidth}
\centering
\includegraphics[width=\linewidth]{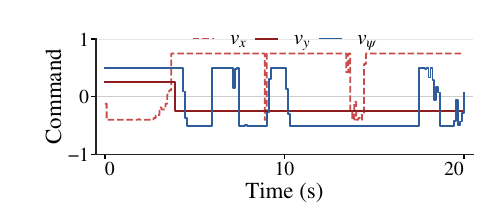}
{\footnotesize (b) Two-Go2 cuboid}
\end{minipage}

\vspace{0.45em}

\begin{minipage}[t]{0.48\linewidth}
\centering
\includegraphics[width=\linewidth]{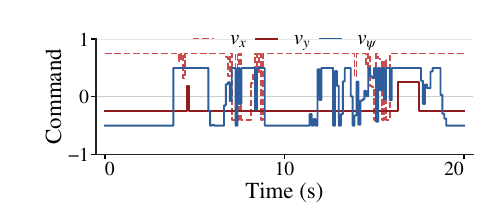}
{\footnotesize (c) Three-Go2 cuboid}
\end{minipage}\hfill
\begin{minipage}[t]{0.48\linewidth}
\centering
\includegraphics[width=\linewidth]{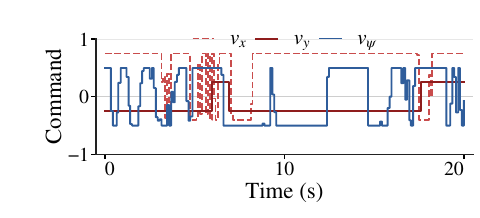}
{\footnotesize (d) Four-Go2 large cuboid}
\end{minipage}
\caption{Post-clamp high-level command waveforms from representative real-robot deployment logs. The plotted channels are the body-frame velocity commands streamed to the Go2 low-level controller.}
\label{fig:appendix_real_robot_command_waveforms}
\end{figure}

\section{Additional Results}
\label{app:additional-results}

\subsection{Semantic-Access Backtest}
\label{app:semantic-backtest}
This subsection reports the semantic-access backtest used in the main comparison.
After the full task stream is completed, old skill parameters are preserved by parameter isolation, but the system still needs to route a pre-execution task query to the appropriate library entry.
The backtest therefore evaluates whether semantic retrieval can recover the intended executable skill after the library has been built.

We construct three English pre-execution retrieval instructions for each of the 14 tasks, giving 42 semantic-access queries in total.
Each query is encoded by the same text-embedding model used by \method{} and matched to the stored skill-description centroids by nearest-neighbor distance.
If the nearest neighbor is the target task's own skill, the query introduces no additional routing error.
If the nearest neighbor is another library skill, we evaluate that selected skill directly on the target task and compare its SR with the target task's own stored skill.

Under the current prompt setting and task stream, semantic access does not create any severe routing failure.
Among the 42 queries, 41 retrieve the target task's own skill, corresponding to a self-retrieval rate of 97.6\%.
The only non-self retrieval occurs for the second rewrite of \texttt{1d-cuboid-rough}, which is routed to \texttt{2d-cuboid-rough}.
For this source-target pair, the target task's own skill reaches 99.2\% average SR, while the retrieved \texttt{2d-cuboid-rough} skill reaches 97.9\% average SR.
Thus, for the current benchmark and prompt interface, semantic access does not change the main conclusion about retained skill availability.

This mismatch also points to the limitation discussed in Sec.~\ref{sec:conclusion}: the nearest semantic neighbor is not necessarily the best zero-shot transfer source, especially when the distance is defined by a prior text embedding model.
The prior semantic distance remains useful, however.
As the skill library grows, methods that select skills with VLM reasoning, rollout trajectories, or empirical success rates must search over an increasing number of entries.
Semantic distance can reduce this search to a smaller candidate set before those more expensive selectors are applied.

\subsection{Semantic Transfer Case-Study Details}
\label{app:case-study-details}
\begin{figure}[t]
\centering
\includegraphics[width=0.96\linewidth]{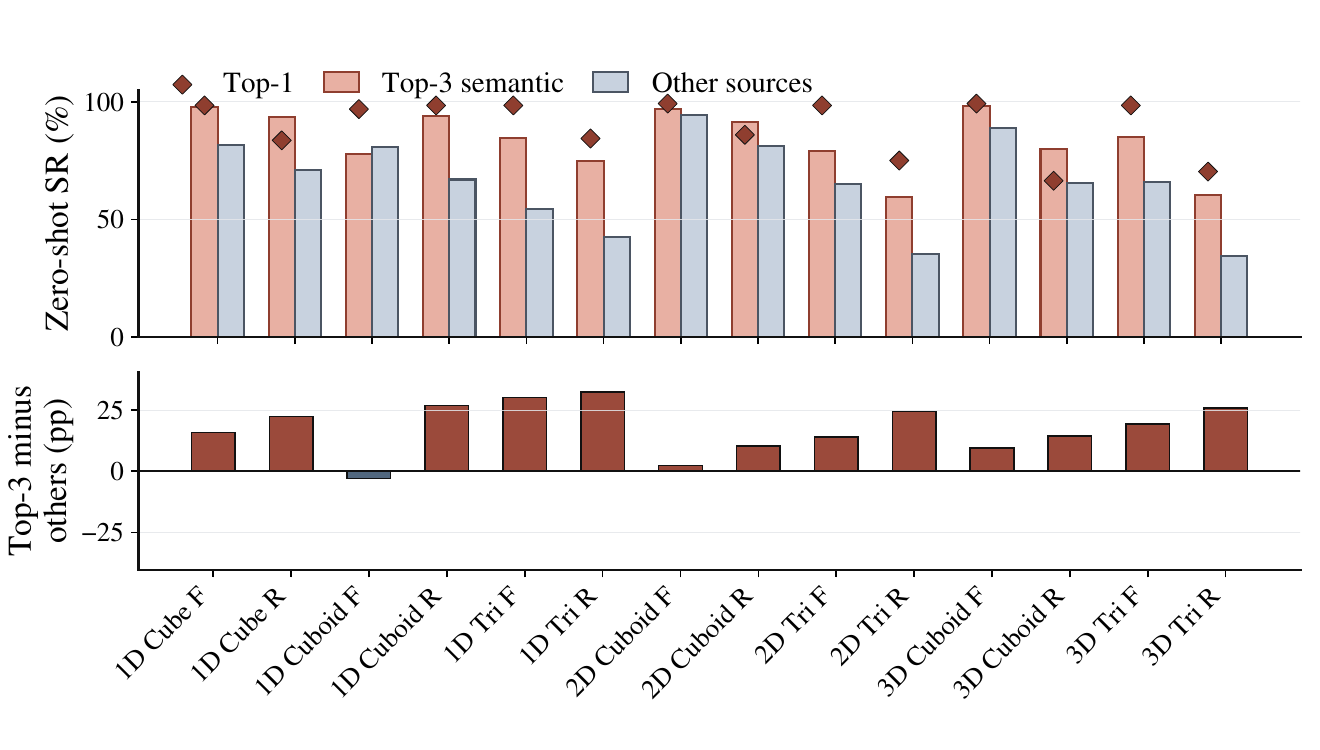}
\caption{Leave-one-target-out semantic transfer summary over all 14 tasks.}
\label{fig:appendix_leave_one_out_summary}
\end{figure}

Following the analysis in Sec.~\ref{sec:ablation-case-analysis}, we further evaluate zero-shot transfer over all 14 tasks to test whether semantic retrieval provides better initialization candidates.
For each target task, we remove the skill trained on that target environment, use Conquer's retrieval rule to select the three nearest semantic source skills, and compare their transfer SR with the average over the remaining source skills.
Figure~\ref{fig:appendix_leave_one_out_summary} shows that the Top-3 semantic sources average 83.8\% SR, while the remaining sources average 66.3\%, supporting the conclusion that semantic similarity improves initialization quality on average.
Table~\ref{tab:appendix_semantic_top3_transfer} gives the detailed source ranking for the \texttt{3d-tri-rough} case visualized in Figure~\ref{fig:semantic_transfer_tsne_3d_tri_rough_c}.
The reported distance is computed between the target task-description query and each source skill descriptor.

\begin{table}[t]
\centering
\caption{Semantic source ranking for \texttt{3d-tri-rough}. Distances are query-to-source descriptor L2 values; SR values are zero-shot source-skill transfer success rates.}
\vspace{8pt}
\label{tab:appendix_semantic_top3_transfer}
\small
\setlength{\tabcolsep}{6pt}
\renewcommand{\arraystretch}{1.08}
\begin{tabular}{@{}clcc@{}}
\toprule
Group & Source skill & Query rank / L2 & Transfer SR (\%) \\
\midrule
Top-3 & \texttt{3d-tri-flat}     & 1 / 0.369 & 70.3 \\
Top-3 & \texttt{3d-cuboid-rough} & 2 / 0.418 & 49.2 \\
Top-3 & \texttt{2d-tri-rough}    & 3 / 0.430 & 60.9 \\
\midrule
Best other & \texttt{2d-tri-flat} & 5 / 0.433 & 68.0 \\
Top-3 mean & -- & ranks 1--3 & 60.2 \\
Other-source mean & -- & ranks 4--13 & 34.4 \\
\bottomrule
\end{tabular}
\end{table}

\subsection{Per-task Final Success Records}
\label{app:per-task-final-sr}
Table~\ref{tab:appendix_per_task_final_sr} reports the final success rate on each benchmark task.
For continual methods, final success is measured after the complete 14-task stream.
For the other baselines, we report their corresponding per-task final evaluation.
Notably, PSEC and Scratch are parameter-isolation baselines that do not include a retrieval module. Accordingly, the final success rates reported in the table are computed under a task-aware selection assumption for these two baselines.
The Backbone column gives zero-shot performance before any task-specific adaptation.

\begin{table}[t]
\centering
\caption{Per-task final SR. Values are percentages.}
\label{tab:appendix_per_task_final_sr}
\small
\setlength{\tabcolsep}{2.6pt}
\renewcommand{\arraystretch}{1.05}
\resizebox{\linewidth}{!}{
\begin{tabular}{@{}lcccccccc@{}}
\toprule
Task & Conquer ($\%$) & Multitask ($\%$) & EWC ($\%$) & Fine-tune ($\%$) & Scratch ($\%$) & PSEC ($\%$) & HiSSD ($\%$) & Backbone ($\%$) \\
\midrule
\texttt{1d-cube-flat}    & 99.5 & 93.2 & 94.9 & 71.1 & 100.0 & 99.2 & 97.7 & 68.0 \\
\texttt{1d-cube-rough}   & 99.5 & 94.5 & 94.9 & 32.8 & 98.4 & 98.4 & 87.5 & 62.5 \\
\texttt{1d-cuboid-flat}  & 99.7 & 100.0 & 94.5 & 32.0 & 100.0 & 99.2 & 100.0 & 77.3 \\
\texttt{1d-cuboid-rough} & 99.5 & 95.1 & 99.2 & 18.0 & 97.7 & 99.2 & 94.5 & 64.8 \\
\texttt{1d-tri-flat}     & 99.5 & 99.5 & 99.2 & 39.8 & 100.0 & 100.0 & 98.4 & 14.8 \\
\texttt{1d-tri-rough}    & 97.1 & 85.9 & 88.7 & 21.9 & 82.0 & 96.1 & 85.2 & 18.8 \\
\texttt{2d-cuboid-flat}  & 100.0 & 100.0 & 98.4 & 84.4 & 99.2 & 100.0 & 100.0 & 99.2 \\
\texttt{2d-cuboid-rough} & 97.7 & 97.1 & 88.3 & 85.2 & 92.2 & 96.1 & 89.8 & 80.5 \\
\texttt{2d-tri-flat}     & 98.7 & 98.2 & 84.4 & 93.0 & 96.1 & 95.3 & 91.4 & 28.1 \\
\texttt{2d-tri-rough}    & 80.2 & 80.7 & 53.1 & 82.8 & 54.7 & 79.7 & 50.0 & 18.8 \\
\texttt{3d-cuboid-flat}  & 96.6 & 97.7 & 96.1 & 98.4 & 99.2 & 97.7 & 98.4 & 95.3 \\
\texttt{3d-cuboid-rough} & 94.0 & 92.7 & 86.3 & 98.4 & 78.9 & 96.1 & 78.1 & 63.3 \\
\texttt{3d-tri-flat}     & 97.9 & 96.1 & 78.5 & 96.1 & 97.7 & 93.8 & 93.8 & 71.9 \\
\texttt{3d-tri-rough}    & 78.4 & 78.9 & 58.2 & 91.4 & 53.1 & 75.0 & 46.1 & 40.6 \\
\bottomrule
\end{tabular}
}
\end{table}

\end{document}